\documentclass{article}

\usepackage{microtype}
\usepackage{graphicx}
\usepackage{subfigure}
\usepackage{booktabs} % for professional tables

\usepackage{hyperref}

\usepackage{hyperref}
\usepackage{url}

\usepackage{array}
\usepackage{color}

\usepackage{multirow}
\usepackage{amssymb}
\usepackage{bm}
\usepackage{balance}
\usepackage{arydshln}
\usepackage{mathtools}
\usepackage{soul}
\usepackage[algo2e]{algorithm2e}
\usepackage{enumitem}
\usepackage{caption}
\usepackage{wrapfig}

\usepackage{booktabs}
\usepackage{amsmath,amsfonts}
\usepackage{xspace}

\usepackage{diagbox}

\usepackage{bbm}

\usepackage{xspace}
\usepackage{textcomp}

\usepackage{tabularx}
\usepackage{verbatim}
\usepackage{tablefootnote}
\usepackage{pifont}

\newcommand{\method}{{Mu$^{2}$}SLAM}
\newcommand{\mslam}{mSLAM}

\newcommand{\xlsrp}{XLS-R}
\newcommand{\xlsrpb}[1]{\xlsrp{} {(#1B)}}

\newcommand{\mtlm}{mSLAM-TLM}
\newcommand{\mtlmb}[1]{\mtlm{} {(#1B)}}
\newcommand{\mctc}{mSLAM-CTC}
\newcommand{\mctcb}[1]{\mctc{} {(#1B)}}

\def\argmin{\mathop{\rm argmin}}%

\def\vx{\mathbf{x}}
\def\vy{\mathbf{y}}
\def\vz{\mathbf{z}}

\def\vm{\mathbf{m}}
\def\vv{\mathbf{v}}

\def\ms{\mathbf{S}}
\def\vs{\mathbf{s}}

\def\vt{\mathbf{t}}

\newcommand{\eat}[1]{} % ignores argument
\newcommand{\eg}{\emph{e.g.}} 
\newcommand{\ie}{\emph{i.e.}}

\newcommand{\checkm}{\ding{52}}
\newcommand{\crossm}{\ding{56}}

\newcommand{\covostmanytomany}{\begin{table*}[!t]
\centering
\begin{tabular}{llllllll}
\toprule
\multirow{2}{*}{Method} &\multirow{2}{*}{\# Encoder} &\multicolumn{4}{c}{xx-en} &en-xx
&{All} \\
\cmidrule(lr){3-6} \cmidrule(lr){7-7} \cmidrule(lr){8-8}
& &High &Med. &Low &Avg &Avg &Avg \\
\midrule
XLS-R &0.3B &30.6 &18.9 &5.1  &13.2 &- &- \\
XLS-R   &1B &34.3 &25.5 &11.7 &19.3 &- &- \\
XLS-R   &2B &36.1 &27.7 &15.1 &22.1 &- &- \\
\midrule
\multicolumn{7}{l}{\it xx-en Multilingual AST\&MT FT} \\
\midrule
mSLAM-TLM &0.6B &35.5 &25.3 &12.3 &19.8  & &- \\
mSLAM-CTC &0.6B &37.6 &27.8 &15.1 &22.4  &- &- \\
mSLAM-CTC &2B &37.8 &29.6 &18.5 &24.8  &- &- \\
Maestro &0.6B &38.2 &31.3 &18.4 &25.2  &- &- \\
Whisper &1.6B &36.2 &32.6 &25.2 &29.1  &- &- \\
\midrule
\midrule
\multicolumn{7}{l}{\it xx-en/en-xx Multilingual AST FT}\\
\midrule
\method{}-char &0.6B &35.0 &28.2 &18.2 &23.8  &28.0 &25.5 \\
\method{}-spm &0.6B &34.4 &27.9 &18.7 &23.9 &27.1 &25.2 \\
\midrule
\multicolumn{7}{l}{\it Multi-task Multilingual FT}\\
\midrule
\method{}-char &0.6B &\bf{37.3} &30.2 &20.5 &26.0  &26.4 &26.2 \\
\method{}-spm &0.6B &37.0 &30.0 &21.2 &26.3 &24.2 &25.4 \\
\midrule
\multicolumn{7}{l}{\it Multi-task Multilingual FT $\rightarrow$ xx-en/en-xx Multilingual AST FT}\\
\midrule
\method{}-char &0.6B &37.0 &30.0 &20.7 &26.0   &\bf{28.4} &27.0 \\
\method{}-spm &0.6B &37.0 &\bf{30.6} &\bf{23.5} &\bf{27.1} &27.9 &\bf{27.4} \\
\bottomrule
\end{tabular}
\caption{Speech translation results on the CoVoST 2 dataset.%We report the average BLEU scores on non-English to English (xx-en), English to non-English (en-xx) and all of translation directions.
}
\label{table:covost_xx2xx}
\end{table*}
}

\newcommand{\covostentoxx}{\begin{table*}[!t]
\centering
\begin{tabular}{lllllll}
\toprule
Model & \# Encoder &En-De &En-Ca &En-Ar &En-Tr &Avg \\
\midrule
wav2vec-2.0  &0.3B &23.8 &32.4 &17.4 &15.4   &22.3 \\
wav2vec-2.0 + LM   &0.3B &24.9 &34.0 &18.0 &16.7 &23.4 \\
SLAM-TLM    &0.6B &27.5 &33.4 &18.9 &16.6  &24.1\\
SLMA-TLM-STM$\rightarrow$w2v-bert &0.6B &27.1 &34.2 &21.2 &17.5 &25.0 \\
SpeechLM-P &0.3B &27.6 &35.9 &21.7 &19.5 &26.2 \\
\midrule
\multicolumn{7}{l}{\it  Multi-task Multilingual FT}\\
\midrule
\method{}-char &0.6B &26.9 &34.2 &20.2 &18.4 &24.9 \\
\method{}-spm. &0.6B &25.1 &32.4 &17.5 &16.1 &22.8 \\
\midrule
\multicolumn{7}{l}{\it Multi-task Multilingual FT $\rightarrow$ Per-language FT}\\
\midrule
\method{}-char &0.6B &\bf{29.4} &\bf{36.1} &\bf{23.3} &\bf{20.4} &\bf{27.3}\\
\method{}-spm. &0.6B &29.1 &\bf{36.1} &22.8  &20.3 &27.1 \\
\bottomrule
\end{tabular}
\caption{Speech translation results on four major English to non-English tasks from the CoVoST 2 dataset.}
\label{table:covost_en2xx}
\end{table*}
}

\newcommand{\covostablation}{\begin{table*}[!t]
\centering
\begin{tabular}{llllllllllll}
\toprule
\multirow{2}{*}{ID} &\multirow{2}{*}{Method} &\multicolumn{3}{c}{Pretrain} &\multicolumn{3}{c}{Finetune}
&\multicolumn{3}{c}{AST} \\
\cmidrule(lr){3-5} \cmidrule(lr){6-8} \cmidrule(lr){9-11}
& &AST &ASR &MT &AST &ASR &MT &xx-en  &en-xx &all \\
\midrule
1 &\method{}-char &\checkm &\crossm &\checkm &\checkm  &\crossm &\crossm &21.1  &24.8 &22.6 \\
2 &\method{}-char &\checkm &\checkm &\crossm &\checkm  &\crossm &\crossm &22.1  &23.7 &22.8 \\
3 &\method{}-char &\crossm &\checkm &\checkm &\checkm  &\crossm &\crossm &22.8  &23.5 &23.1 \\
4 &\method{}-char &\checkm &\checkm &\checkm &\checkm  &\crossm &\crossm &23.1  &24.1 &23.5 \\
\midrule
5 &\method{}-char &\checkm &\crossm &\checkm &\checkm  &\checkm &\checkm &23.8  &25.1 &24.4 \\
6 &\method{}-char &\checkm &\checkm &\crossm &\checkm  &\checkm &\checkm &25.4  &24.5 &25.0 \\
7 &\method{}-char &\crossm &\checkm &\checkm &\checkm  &\checkm &\checkm &24.5  &22.8 &23.8 \\
8 &\method{}-char &\checkm &\checkm &\checkm &\checkm  &\checkm &\checkm &24.7  &24.4 &24.6 \\
\bottomrule
\end{tabular}
\caption{Effect of paired data when being incorporated into pretraining or finetuning stages.
}
\label{table:covost_ablation}
\end{table*}
}

\newcommand{\covostperlanguage}{
\begin{table*}[hbt!]
\centering
\caption{BLEU results on xx-en CoVoST.}
\label{tab:covost_xxen_per}
\begin{tabular}{llllllllllllll}
\toprule
& \multicolumn{4}{c}{High-resource} & \multicolumn{5}{c}{Mid-resource} & \multicolumn{3}{c}{Low-resource}\\
\cmidrule(lr){2-5} \cmidrule(lr){6-10} \cmidrule(lr){11-13}
xx-en & fr & de & es & ca & fa & it & ru & pt & zh & tr & ar & et \\
Train Hours & 264h & 184h & 113h & 136h & 49h & 44h & 18h & 10h & 10h & 4h & 2h & 3h \\
\midrule
\xlsrpb{0.3} & 32.9 & 26.7 & 34.1 & 28.7 & 5.9 & 29.0 & 26.4 & 28.3 & 4.9 & 4.6 & 3.0 & 3.5 \\
\xlsrpb{1} & 36.2 & 31.2 & 37.9 & 31.9 & 9.6  & 33.1 & 37.0 & 39.3 & 8.7 & 12.8 & 12.2 & 8.3 \\
\xlsrpb{2} &   37.6 &   33.6 &   39.2 &   33.8 &   12.9 &   34.9 & 39.5 &   41.8 &   9.4 &   16.7 &   17.1 & 11.1 \\
\midrule
\multicolumn{8}{l}{xx-en Multilingual AST\&MT FT} \\
\midrule
\mtlmb{0.6} & 36.8 & 32.8 & 38.8 & 33.6 & 9.7 & 34.6 & 41.2 & 32.1 &  8.8 & 12.2 & 12.6 & 16.6 \\
\mctcb{0.6} & 38.6 & 36.1 & 40.6 & 35.2 & 7.2 & 37.0 & 47.5 & 36.4 & 10.8 & 15.6 & 14.2 & 20.3 \\
\mctcb{2}   & 39.0 & 35.9 & 41.0 & 35.4	& 9.7 & 37.3 & 48.4	& 42.8 & 10.0 & 24.2 & 19.3	& 22.6	\\
\midrule
\multicolumn{8}{l}{\it xx-en/en-xx Multilingual AST FT}\\
\midrule
\method{}-char(0.6B)  & 36.5 & 31.3 & 38.1 & 34.0 & 12.2 & 33.8 & 40.0 & 40.3 & 14.5 & 21.2 & 25.6 & 14.0 \\
\method{}-spm (0.6B)  & 35.6 & 30.8 & 37.9 & 33.1 & 11.1 & 33.3 & 40.0 & 40.4 & 14.8 & 20.7 & 27.3 & 12.8 \\
\midrule
\multicolumn{8}{l}{\it Multi-task Multilingual FT}\\
\midrule
\method{}-char(0.6B) & 39.1 & 34.9 & 40.6 & 34.6 & 14.7 & 34.7 & 42.7 & 43.7 & 15.3 & 23.5 & 26.3 & 14.8 \\
\method{}-spm (0.6B) & 38.4 & 34.7 & 40.5 & 34.2 & 14.9 & 35.0 & 41.2 & 43.2 & 15.8 & 23.3 & 29.5 & 14.9 \\
\midrule
\multicolumn{8}{l}{\it Multi-task Multilingual FT $\rightarrow$ xx-en/en-xx Multilingual AST FT}\\
\midrule
\method{}-char(0.6B)  & 38.7 & 35.0 & 40.2 & 34.2 & 14.5 & 34.2 & 43.2 & 43.3 & 15.2 & 23.6 & 27.2 & 14.6 \\
\method{}-spm (0.6B)  & 38.5 & 34.9 & 40.6 & 34.4 & 15.0 & 34.9 & 43.4 & 44.0 & 16.3 & 23.9 & 31.4 & 15.4 \\
\bottomrule
\end{tabular}    
\begin{tabular}{llllllllllllll}
\toprule
 & \multicolumn{9}{c}{Low-resource} & \multicolumn{4}{c}{Average}\\
\cmidrule(lr){2-10} \cmidrule(lr){11-14}
xx-en &  mn & nl & sv & lv & sl & ta & ja & id & cy & high & mid & low & all \\
Train Hours & 3h & 7h & 2h & 2h & 2h & 2h & 2h & 2h & 2h	& \\
\midrule 
\xlsrpb{0.3}  & 0.4 & 22.0 & 10.3 & 6.0 & 6.6 & 0.2 & 0.6 & 1.4 & 2.5 & 30.6 & 18.9 & 5.1 & 13.2 \\
\xlsrpb{1}  & 0.8 & 28.2 & 24.7 & 16.0 & 16.7 & 0.3 & 1.9 & 10.3 &  8.6 & 34.3 & 25.5 & 11.7 & 19.3 \\
\xlsrpb{2}  &   1.6 &   31.7 &   29.6 &   19.5 &   19.6 & 0.5 &   3.5 &   16.5 &   14.0 &   36.1 &   27.7 &   15.1 &   22.1 \\
\midrule
\multicolumn{8}{l}{xx-en Multilingual AST\&MT FT} \\
\midrule
\mtlmb{0.6} & 0.3 & 33.2 & 26.3 & 15.2 & 19.8 & 0.5 & 1.3 & 3.7 & 5.6 & 35.5 & 25.3 & 12.3 & 19.8 \\
\mctcb{0.6} & 0.9 & 36.3 & 31.7 & 19.8 & 25.6 & 0.5 & 2.4 & 6.1 & 7.7 & 37.6 & 27.8 & 15.1 & 22.4  \\
\mctcb{2} & 0.8 & 37.6 & 38.5	& 26.8 & 32.3 & 0.6	& 3.3 & 8.8	& 6.7 & 37.8 & 29.6	& 18.5 & 24.8 \\
\midrule
\multicolumn{8}{l}{\it xx-en/en-xx Multilingual AST FT}\\
\midrule
\method{}-char(0.6B) & 2.8 & 27.0 & 33.0 & 20.1 & 21.7 & 1.4 & 7.3 & 27.1 & 17.3 & 35.0 & 28.2 & 18.2 & 23.8\\
\method{}-spm(0.6B)  & 2.7 & 26.8 & 30.6 & 19.7 & 24.3 & 2.4 & 9.1 & 29.4 & 18.9 &34.4 &27.9 &18.7 &23.9 \\
\midrule
\multicolumn{8}{l}{\it Multi-task Multilingual FT}\\
\midrule
\method{}-char(0.6B)  & 4.0 & 31.6 & 34.6 & 20.7 & 26.0 & 1.9 & 9.0 & 29.2 & 24.2 &37.3 &30.2 &20.5 &26.0 \\
\method{}-spm (0.6B)  & 4.3 & 31.6 & 31.5 & 20.5 & 25.7 & 2.3 & 11.0 & 33.8 & 26.2 &37.0 &30.0 &21.2 &26.3 \\
\midrule
\multicolumn{8}{l}{\it Multi-task Multilingual FT $\rightarrow$ xx-en/en-xx Multilingual AST FT}\\
\midrule
\method{}-char(0.6B)  & 4.1 & 31.6 & 34.2 & 21.1 & 25.6 & 2.2 & 9.1 & 30.8 & 24.0 &37.0 &30.0 &20.7 &26.0  \\
\method{}-spm (0.6B)  & 4.5 & 32.1 & 34.8 & 20.4 & 27.5 & 2.5 & 11.8 & 36.1 & 27.3  &37.0 &30.6 &23.5 &27.1 \\
\bottomrule

\end{tabular} 
\end{table*}
}

\newcommand{\covostperlanguageentoxx}{
\begin{table*}[hbt!]
\centering
\caption{BLEU results on en-xx CoVoST.}
\label{tab:covost_enxx_per}
\begin{tabular}{llllllllllllll}
\toprule
en-xx &ar &ca &cy &de &et &fa &id &ja &lv &mn &sl &sv \\
Train Hours & 430h & 430h & 430h & 430h & 430h & 430h & 430h & 430h & 430h & 430h & 430h & 430h \\
\midrule
\multicolumn{8}{l}{\it xx-en/en-xx Multilingual AST FT}\\
\midrule
\method{}-char(0.6B)   & 22.5 & 35.7 & 37.0 & 28.6 & 24.6 & 20.1 & 33.1 & 32.2 & 24.9 & 17.6 & 30.1 & 35.6 \\
\method{}-spm (0.6B)   & 21.2 & 35.0 & 35.3 & 27.7 & 23.5 & 20.2 & 33.0 & 32.0 & 23.4 & 16.8 & 28.0 & 34.4 \\
\midrule
\multicolumn{8}{l}{\it Multi-task Multilingual FT}\\
\midrule
\method{}-char(0.6B)  & 20.2 & 34.2 & 35.9 & 26.9 & 23.1 & 19.2 & 31.8 & 30.9 & 23.0 & 16.3 & 27.8 & 34.5  \\
\method{}-spm (0.6B)  & 17.5 & 32.4 & 32.8 & 25.1 & 20.8 & 18.9 & 31.0 & 28.9 & 19.6 & 14.6 & 23.4 & 32.0 \\
\midrule
\multicolumn{8}{l}{\it Multi-task Multilingual FT $\rightarrow$ xx-en/en-xx Multilingual AST FT}\\
\midrule
\method{}-char(0.6B)  & 22.9 & 36.3 & 37.4 & 29.0 & 25.1 & 20.2 & 33.5 & 32.5 & 25.5 & 17.9 & 30.7 & 36.2  \\
\method{}-spm (0.6B)  & 22.2 & 35.6 & 36.6 & 28.7 & 24.2 & 20.6 & 33.8 & 32.1 & 24.4 & 17.2 & 29.4 & 35.5 \\
\bottomrule
\end{tabular}    
\begin{tabular}{lllll}
\toprule
en-xx &ta &tr &zh & all \\
Train Hours & 430h & 430h & 430h & 430h \\
\midrule
\multicolumn{5}{l}{\it xx-en/en-xx Multilingual AST FT}\\
\midrule
\method{}-char(0.6B)  & 21.2 & 19.9 & 37.4 &28.0  \\
\method{}-spm(0.6B)   & 20.5 & 18.7 & 36.8 &27.1\\
\midrule
\multicolumn{5}{l}{\it Multi-task Multilingual FT}\\
\midrule
\method{}-char(0.6B)  & 19.4 & 18.4 & 34.9  &26.4\\
\method{}-spm (0.6B)   & 17.7 & 16.1 & 32.9 &24.2\\
\midrule
\multicolumn{5}{l}{\it Multi-task Multilingual FT $\rightarrow$ xx-en/en-xx Multilingual AST FT}\\
\midrule
\method{}-char(0.6B)  & 20.9 & 20.1 & 38.1  &28.4 \\
\method{}-spm (0.6B)  & 20.9 & 19.4 & 37.4  &27.9 \\
\bottomrule

\end{tabular} 
\end{table*}
}

\newcommand{\asrvp}{
\begin{table}[!t]
\small
\centering
\begin{tabular}{ll}
\toprule
Method &WER \\
\midrule
\em{Transducer as Decoder} \\
\midrule
XLS-R (0.3B)  &12.8 \\
XLS-R (1B)    &10.6  \\
mSLAM-TLM (0.6B)  &9.4  \\
mSLAM-CTC (0.6B)  &9.2  \\
mSLAM-CTC (2B)  &9.1 \\
Maestro (0.6B)  &8.1  \\
Whisper (1.6B)  &13.6  \\
\midrule
\multicolumn{2}{l}{\it Transformer as Decoder, ASR Multilingual FT} \\
\midrule
\method{}-char (0.7B)  &9.8  \\
\method{}-spm (0.7B)  &9.5   \\
\midrule
\multicolumn{2}{l}{\it Transformer as Decoder, Multi-task Multilingual FT} \\
\midrule
\method{}-char (0.7B)  &31.5  \\
\method{}-spm (0.7B)  &32.5   \\
\midrule
\multicolumn{2}{l}{\it Transformer as Decoder,} \\
\multicolumn{2}{l}{\it Multi-task Multilingual FT $\rightarrow$ ASR multilingual FT} \\
\midrule
\method{}-char (0.7B)  &9.7  \\
\method{}-spm (0.7B)  &9.2   \\
\bottomrule
\end{tabular}
\caption{Speech recognition results on the VoxPopuli dataset. %Average Word Error Rate (WER) are used to evaluate ASR outputs.
}
\label{table:asrvp}
\end{table}
}

\newcommand{\vpperlanguage}{
\begin{table*}[bht!]
\centering
\caption{VoxPopuli ASR results in terms of WER.}
\begin{tabular}{lllllllll}
\toprule
& en & de & it & fr & es & pl & ro & hu  \\
\midrule
Train Hours & 543h & 282h & 91h & 211h & 166h & 111h & 89h & 63h \\
\midrule 
\xlsrpb{0.3} & 10.2 & 13.0 & 19.2 & 12.6 & 9.8 & 9.6 & 7.9 & 11.6 \\
\xlsrpb{1}  &    8.8 &   11.5 &   15.1 &   10.8 &   8.2 &   7.7 &   7.3 &   9.6 \\
\mtlmb{0.6} & 7.3 & 8.9 & 15.6 & 9.3 & 8.7 & 6.5 & 8.5 & 8.4 \\
\mctcb{0.6} & 7.1 & 8.9 & 15.6 & 9.3 & 8.6 & 6.5 & 8.5 & 8.1  \\
\mctcb{2} & 7.0	  & 8.7 & 15.4 & 9.4 & 8.4 & 6.4 & 7.8 & 8.4 \\
\midrule
\multicolumn{9}{l}{\it Transformer as Decoder, ASR Multilingual FT} \\
\midrule
\method{}-char (0.7B)    & 8.0 & 10.2 & 16.4 & 9.7 & 9.1 & 7.0 & 8.0 & 9.0 \\
\method{}-spm (0.7B)    & 7.5 & 8.9 & 14.4 & 9.1 & 7.9 & 7.1 & 7.5 & 8.9  \\
\midrule
\multicolumn{9}{l}{\it Transformer as Decoder, Multi-task Multilingual FT} \\
\midrule
\method{}-char (0.7B)   & 28.1 & 29.4 & 48.5 & 32.2 & 36.0 & 29.2 & 32.5 & 31.5  \\
\method{}-spm (0.7B)   & 28.2 & 29.7 & 49.5 & 32.4 & 36.3 & 29.7 & 32.9 & 32.5  \\
\midrule
\multicolumn{9}{l}{\it Transformer as Decoder, Multi-task Multilingual FT $\rightarrow$ ASR multilingual FT} \\
\midrule
\method{}-char (0.7B)  & 7.8 & 9.5 & 16.4 & 9.4 & 9.2 & 7.0 & 8.1 & 8.8
 \\
\method{}-spm (0.7B)   & 7.2 & 8.5 & 13.9 & 8.6 & 7.5 & 6.8 & 7.1 & 8.7   \\
\bottomrule
\toprule
& nl & cs & sl & fi & hr & sk & Avg \\
\midrule
Labeled data & 53h & 62h & 10h & 27h & 43h & 35h & \\
\midrule 
\multicolumn{8}{l}{\it Prior work \citep{babu2021xls}} \\
\midrule
\xlsrpb{0.3} & 14.8 & 10.5 & 24.5 & 14.2 & 12.3 & 8.9 & 12.8\\
\xlsrpb{1}  &   12.5 &   8.7 &   19.5 &   11.3 &   10.0 &   7.1 &   10.6 \\
\mtlmb{0.6} & 10.5 & 7.1 & 15.8 & 9.0 & 10.0 & 6.2 & 9.4\\
\mctcb{0.6} & 10.3 & 7.0 & 14.2 & 9.2 & 9.1 & 5.9 & 9.2\\
\mctcb{2} &10.5	& 6.8 & 15.1 & 8.7 & 9.1 & 6.0 & \textbf{9.1}\\
\midrule
\multicolumn{9}{l}{\it Transformer as Decoder, ASR Multilingual FT} \\
\midrule
\method{}-char (0.7B)  & 11.3 & 7.7 & 15.0 & 10.1 & 8.9 & 6.4  &9.8  \\
\method{}-spm (0.7B)  & 11.4 & 7.6 & 17.1 & 10.5 & 8.8 & 6.4 &9.5   \\
\midrule
\multicolumn{9}{l}{\it Transformer as Decoder, Multi-task Multilingual FT} \\
\midrule
\method{}-char (0.7B)  & 28.1 & 20.0 & 34.6 & 31.9 & 39.2 & 29.3 &32.8  \\
\method{}-spm (0.7B)  & 28.4 & 29.5 & 35.8 & 32.8 & 39.8 & 29.9 &33.4   \\
\midrule 
\multicolumn{9}{l}{\it Transformer as Decoder, Multi-task Multilingual FT $\rightarrow$ ASR multilingual FT} \\
\midrule
\method{}-char (0.7B)   & 11.6 & 7.1 & 16.3 & 10.3 & 8.5 & 6.1
 &9.7  \\
\method{}-spm (0.7B)   & 10.9 & 7.4 & 16.3 & 10.7 & 8.6 & 6.5 &9.2   \\
\bottomrule
\end{tabular}    
\label{tab:vp_asr}
\end{table*}
}

\newcommand{\xtremexnli}{\begin{table}[!t]
\centering
\small
\begin{tabular}{lllll}
\toprule
Model & En &Eu. &Non-Eu. &Avg \\
\midrule
\multicolumn{5}{l}{\it Zero-shot} \\
\midrule
mT5-Small (0.3B)  &79.6 &66.6 &60.4 &63.8 \\
mT5-Base (0.6B)  &84.5 &77.1 &69.5 &73.0 \\
mSLAM (0.6B)  &80.4 &71.4 &49.5 &58.9 \\
mSLAM (2B) &80.1 &74.4 &59.9 &66.1 \\
\midrule
\method{}-char (0.7B) &76.5 &65.9 &56.6 &60.9 \\
\method{}-spm (0.7B) &81.2 &71.9 &61.6 &\bf{66.4} \\
\midrule
\multicolumn{5}{l}{\it Translate-Train-All} \\
\midrule
mT5-Small (0.3B)  &78.3 &73.6 &69.2 &71.3 \\
mT5-Base (0.6B)  &85.9 &82.1 &77.9 &79.8 \\
mSLAM (0.6B)  &81.1 &76.0 &65.5 &70.0 \\
mSLAM (2B) &84.1 &80.5 &73.7 &76.1 \\
\midrule
\method{}-char (0.7B) &79.0 &75.5 &70.6 &72.9 \\
\method{}-spm (0.7B) &83.3 &78.8 &73.8 &\bf{76.1} \\
\bottomrule
\end{tabular}
\caption{Text classification results on the validation sets in XNLI.
}
\label{table:xnli}
\end{table}
}

\newcommand{\xtremeqa}{\begin{table*}[!t]
\centering
\begin{tabular}{llll}
\toprule
Method  & English &Non-English &Avg\\
\midrule
\multicolumn{4}{l}{\it Zero-shot} \\
\midrule
mT5-Small (0.3B)  &53.9/43.6 &32.6/20.9 &35.2/23.2 \\
mT5-Base (0.6B)  &71.8/60.9 &56.4/41.8  &57.2/41.2 \\
\midrule
\method{}-char (0.7B) &56.1/47.0 &20.9/13.9  &25.0/18.0 \\
\method{}-spm (0.7B) &\textbf{59.6}/\textbf{47.7} &\textbf{22.1}/\textbf{14.6}  &\textbf{26.6}/\textbf{18.7} \\
\midrule
\multicolumn{4}{l}{\it Translate-Train-All} \\
\midrule
mT5-Small (0.3B)  &57.1/46.6 &47.1/32.2 &48.2/34.0 \\
mT5-Base (0.6B)  &71.1/58.9 &63.2/46.4  &64.0/47.7 \\
\midrule
\method{}-char (0.7B) &62.1/53.0 &53.5/\bf{41.6}  &54.3/\bf{42.8} \\
\method{}-spm (0.7B) &\bf{67.9}/56.1 &\textbf{54.5}/40.6 &\textbf{55.9}/42.3 \\
\bottomrule
\end{tabular}
\caption{TyDiQA-GoldP results (F1/EM) on the test sets.}
\label{table:tydiqa}
\end{table*}
}

\newcommand{\xnliperlanguage}{
\begin{table*}[bht!]
        \begin{center}
            \caption{XNLI dev accuracy for all 15 languages.}
            \resizebox{0.95\linewidth}{!}{
            \begin{tabular}[h!]{lllllllllllllllll}
            \toprule
                {\bf Model} & \bf{en} &	\bf{ar} &	\bf{bg} &	\bf{de} &	\bf{el} &	\bf{es} &	\bf{fr} &	\bf{hi} &	\bf{ru} &	\bf{sw} &	\bf{th} &	\bf{tr} &	\bf{ur} &	\bf{vi} &	\bf{zh} &	\bf{Avg} \\
                \midrule
                mT5-Small (0.3B) & 79.6 & 62.2 & 67.8 &	64.8 & 65.8 & 68.4 & 66.2 & 59.0 & 65.3 &	55.4 & 63.2 & 58.9 & 54.5 & 61.8 & 63.4 & 63.8\\
                mT5-Base (0.6B) & 84.5 & 71.2 & 76.9 &	75.6 & 76.3 & 79.0 & 77.7 & 66.9 & 74.9 &	63.6 & 70.0 & 69.2 & 64.8 & 72.0 & 72.5 & 73.0\\
                \midrule
                \multicolumn{9}{l}{\it Zero-shot} \\
                \midrule
                \mtlmb{0.6} & 75.7&47.3&56.7&	55.1	& 52.2&60.9&62.8&48.6&	58.5& 	46.0&	46.9&	51.3&	47.2&	50.7&41.0 & 53.4\\
                \mctcb{0.6}& 80.4 & 46.5 & 69.8&72.1&	67.5&74.7 & 72.9 & 42.0 & 68.7 & 45.5 &	42.9 & 48.7 & 44.2 & 63.3 &	43.3  & 58.9\\
                \mctcb{2} & 80.1 & 61.1 & 73.3 & 74.7 & 72.7 & 76.0 & 75.3 & 59.4 & 70.9 & 52.2 & 56.8 & 63.9 & 59.0 & 65.9 & 50.1 & 66.1\\
                \midrule
                \method{}-char (0.7B) & 76.5 & 60.6 & 62.1 & 64.1 & 62.6 & 68.0 & 66.4 & 58.3 & 61.7 & 44.4 & 55.8 & 58.4 & 55.4 & 60.0 & 59.6 &60.9 \\
                \method{}-spm (0.7B)  & 81.2 & 65.7 & 67.4 & 71.4 & 65.7 & 74.1 & 74.2 & 57.2 & 69.2 & 51.1 & 63.0 & 63.9 & 55.1 & 66.0 & 70.8 &66.4 \\
                \midrule
                \multicolumn{9}{l}{\it Translate-Train-All} \\
                \midrule
                mT5-Small (0.3B) & 78.3 & 68.8 & 73.5 &	73.2 & 73.4 & 74.4 & 73.5 & 67.4 & 71.1 &	67.2 & 71.1 & 69.9 & 63.6 & 70.5 & 72.9 & 71.3\\
                mT5-Base (0.6B) & 85.9 & 78.8 & 82.2 &	81.6 & 81.4 & 83.0 & 82.1 & 77.0 & 81.1 &	74.8 & 78.6 & 78.4 & 73.3 & 78.9 & 80.2 & 79.8\\
                \midrule
                \mtlmb{0.6} & 74.3& 64.2 & 68.7 & 69.5 & 69.2 &	70.2 &	71.4 &	64.5 &	65.4 &	63.4 &	65.6 &	65.9 &	62.4 &	67.3 &	64.4 & 67.1 \\
                \mctcb{0.6} & 81.1&63.5&76.7&76.0&73.1&77.8&76.4&63.6&73.1&64.1&64.9&66.8&60.5&68.4&64.5 & 70.0 \\
                \mctcb{2} & 84.1 &	80.2 &	80.1 &	78.7 &	82.9 &	80.5 &	74.4 &	72.1 &	76.8 &	71.7 &	73.8 &	76.2 &	69.8 &	75.9 &	72.8 & 76.1 \\
                \midrule
                \method{}-char (0.7B)  & 79.0 & 72.0 & 75.0 & 74.0 & 75.2 & 77.7 & 75.4 & 70.1 & 72.4 & 69.7 & 70.9 & 72.1 & 65.7 & 73.2 & 71.2 &72.9 \\
                \method{}-spm (0.7B)  & 83.3 & 74.6 & 77.0 & 77.9 & 77.6 & 80.6 & 79.2 & 71.9 & 76.2 & 71.3 & 73.1 & 76.7 & 67.8 & 76.5 & 78.5 &76.1 \\
                \bottomrule
            \end{tabular}
            }
       \vspace{-0.4cm}
        \end{center}
    \end{table*}
}

\newcommand{\tydiqaperlanguage}{
\begin{table*}[bht!]
        \begin{center}
            \caption{TyDiQA GoldP test results (F1/EM) for all 9 languages.}
            \resizebox{0.95\linewidth}{!}{
            \begin{tabular}[h!]{lllllllllll}
            \toprule
                {\bf Model} & \bf{ar} &	\bf{bn} &	\bf{en} &	\bf{fi} &	\bf{id} &	\bf{ko} &	\bf{ru} &	\bf{sw} &	\bf{te} &	\bf{avg} \\
                \midrule
                 \multicolumn{9}{l}{\it Zero-shot} \\
                \midrule
                mT5-Small (0.3B) &41.1/26.0 &18.9/13.3 &53.9/43.6 &39.2/22.6 &44.4/31.7 &24.9/16.3 &40.5/24.3 &34.8/21.2 &16.9/11.5 &34.9/23.4 \\
                mT5-Base (0.6B)  &67.1/50.4 &40.7/22.1 &71.8/60.9 &67.0/52.2 &71.3/54.5 &49.5/37.7 &54.9/32.6 &60.4/43.9 &40.6/31.1 & 58.1/42.8\\
                \midrule
                \method{}-char (0.7B)   & 10.8/3.4 & 1.8/1.8 & 58.1/50.2 & 31.6/21.9 & 37.7/24.8 & 12.3/9.8 & 23.1/12.1 & 37.0/27.1 & 13.4/10.8 & 25.1/18.0 \\
                \method{}-spm (0.7B)  & 17.9/9.3 & 1.3/0.9 & 62.3/51.4 & 33.5/22.0 & 39.3/25.3 & 7.2/6.2 & 26.1/15.8 & 36.5/25.6 & 15.3/11.5 & 26.6/18.7 \\
                \midrule
                \multicolumn{9}{l}{\it Translate-Train-All} \\
                \midrule
                mT5-Small (0.3B) &56.8/39.7 &37.2/21/2 &57.1/46.6 &50.9/37.2 &60.1/45.1 &40.4/29.3 &50.7/33.6 &51.5/35.3 &29.3/18.1 &48.2/34.0 \\
                mT5-Base (0.6B) &68.0/50.2 &57.4/35.4 &71.1/58.9 &68.8/55.2 &73.5/57.2 &56.5/43.8 &64.0/45.8 &65.8/48.3 &51.2/34.1 &64.0/47.7  \\
                \midrule
                \method{}-char (0.7B)  & 60.1/41.8 & 46.6/35.4 & 62.1/53.0 & 58.5/47.3 & 55.4/40.5 & 51.5/45.3 & 52.4/36.1 & 67.0/57.1 & 36.2/29.0 & 54.4/42.8  \\
                \method{}-spm (0.7B)  & 61.8/42.8 & 46.0/32.7 & 67.3/55.9 & 61.8/47.8 & 66.3/48.0 & 48.0/39.5 & 53.9/33.9 & 69.8/61.1 & 28.1/18.8 & 55.9/42.3 \\
                \bottomrule
            \end{tabular}
            }
       \vspace{-0.4cm}
        \end{center}
    \end{table*}
}

\usepackage[accepted]{icml2023}

\usepackage{amsmath}
\usepackage{amssymb}
\usepackage{mathtools}
\usepackage{amsthm}

\usepackage[capitalize,noabbrev]{cleveref}

\theoremstyle{plain}

\theoremstyle{definition}

\theoremstyle{remark}

\usepackage[textsize=tiny]{todonotes}

\icmltitlerunning{Multitask, Multilingual Speech and Language Models}

\begin{document}

\twocolumn[
\icmltitle{\method: Multitask, Multilingual Speech and Language Models}




\begin{icmlauthorlist}
\icmlauthor{Yong Cheng}{google}
\icmlauthor{Yu Zhang}{google}
\icmlauthor{Melvin Johnson}{google}
\icmlauthor{Wolfgang Macherey}{google}
\icmlauthor{Ankur Bapna}{google}
\end{icmlauthorlist}

\icmlaffiliation{google}{Google Research, Google LLC, USA}

\icmlcorrespondingauthor{Yong Cheng}{chengyong@google.com}

\icmlkeywords{Machine Learning, ICML}

\vskip 0.3in
]



\printAffiliationsAndNotice{}  % leave blank if no need to mention equal contribution

\begin{abstract}
We present \method{}, a multilingual sequence-to-sequence model pre-trained jointly on unlabeled speech, unlabeled text and supervised data spanning Automatic Speech Recognition (ASR), Automatic Speech Translation (AST) and Machine Translation (MT), in over 100 languages. By leveraging a quantized representation of speech as a target, \method{} trains the speech-text models with a sequence-to-sequence masked denoising objective similar to T5 on the decoder and a masked language modeling objective (MLM) on the encoder, for both unlabeled speech and text, while utilizing the supervised tasks to improve cross-lingual and cross-modal representation alignment within the model. On CoVoST AST, \method{} establishes a new state-of-the-art for models trained on public datasets, improving on xx-en translation over the previous best by 1.9 BLEU points and on en-xx translation by 1.1 BLEU points. On Voxpopuli ASR, our model matches the performance of an \mslam{} model fine-tuned with an RNN-T decoder, despite using a relatively weaker Transformer decoder. On text understanding tasks, our model improves by more than 6\% over \mslam{} on XNLI, getting closer to the performance of mT5 models of comparable capacity on XNLI and TydiQA, paving the way towards a single model for all speech and text understanding tasks.
\end{abstract}

\section{Introduction}

% Version 0.1
% The tremendous success of self-supervised learning of multilingual text representations~\cite{devlin2018multilingualbert,liu2020multilingual,conneau2019unsupervised,xue2020mt5,hu2020xtreme} incentivizes the speech research community to move forward to the multilingual modeling of speech like XLSR~\cite{}
% % finetuning
% Our pre-training method inherits the idea of BERT~\cite{devlin2018bert} to reconstruct the masked tokens according to the context unmasked tokens. However, the artificial token [MASK] used in pre-training are absent from label data in fine-tuning. The discrepancy between pre-training and fine-tuning hinders the model from being adequately optimized on the downstream applications. To alleviate this issue, we propose a gradual fine-tuning by 

% Version 1.0 notes

The recent rapid developments in NLP have witnessed the tremendous success of moving towards unified text models for both understanding and generation tasks across hundreds of languages, evolving into numerous pre-trained models %based on different architectures,
from encoder-only models focusing on text understanding ~\cite{devlin2018bert,devlin2018multilingualbert}, to decoder-only models~\cite{radford2018improving,chowdhery2022palm} and encoder-decoder models~\cite{song2019mass,lewis2019bart,raffel2020exploring,xue2020mt5} for both understanding and generation. The speech pre-training methods have shown a similar trend towards unified models from the dominant encoder-only models~\cite{baevski2020wav2vec,hsu2021hubert,babu2021xls,bapna:2021,bapna:2022} to generative models on cross-modal speech and text data,  exemplified by a couple of recent trails such as decoder-only models~\cite{borsos2022audiolm} and encoder-decoder models~\citep{ao2021speecht5,chen2022maestro,sainath2022joist,Zhou2022MMSpeechMM,zhang2022speechut,tang2022unified}.

Although these works have achieved impressive performance, they only consider partial aspects of the unified models in speech and text. First, except for SLAM and mSLAM~\cite{bapna:2021,bapna:2022}, most of them merely focus on speech-related tasks by taking text data as auxiliary inputs while ignoring evaluations on text-related benchmarks, which leaves us unknown to gauge the effect of interference and capacity dilution. Second, there are few studies investigating multilingual modeling with both speech and text~\cite{bapna:2022,chen2022maestro}, which limits them in leveraging cross-lingual transfer to enrich the speech and text joint representations. Third, multi-task learning has demonstrated the effectiveness of inductive transfer to improve model generalization, yet it is understudied in speech-text pre-training~\cite{tang2022unified,zhang2022speechut,chen2022maestro} where they explicitly differentiate the utilization of labeled data in pre-training by introducing customized networks and losses. 
%The most related Maestro~\cite{chen2022maestro} attempts to apply multi-task pretraining on AST, ASR and MT data too, but it highly depends on a pre-trained~\mslam{} and the pre-trained decoder is tailored to ASR.
Fourth, it is essential for prior speech-text models to design modality-specific blocks and losses to yield high performance~\cite{bapna:2022,chen2022maestro,tang2022unified,zhang2022speechut,zhang2022speechlm} which somewhat violates the principle of the unified models by using one model for all tasks, thus undermining the language and modality transfer to learn general speech-text shared representations.

In this work, we propose a multi-task multilingual pre-training method based on an encoder-decoder model, called~\method{}. The speech-text model is jointly pre-trained on a set of different tasks involving unlabeled speech, unlabeled text, labeled speech-text (ASR\&AST), and labeled text-text (MT). We scale up the language type in both speech and text to more than 100, covering the majority of mainstream spoken languages. 
For the simplicity of extending our current pre-training to more data, we unify the pre-training losses for unlabeled and labeled data by defining a masked language modeling (MLM) loss on the encoder~\cite{devlin2018bert}, a similar T5 loss on decoder~\cite{song2019mass,raffel2020exploring} and an alignment loss only for the labeled data. To enforce the sharing and take full advantage of modality capacity for speech and text, we minimize the number of modality-specific layers in our model design with only a conventional CNN block used to extract speech representations, which pushes forward speech-text models towards the unified models. As our pre-training method inherits the idea of BERT~\cite{devlin2018bert} to reconstruct the masked tokens according to the contextual unmasked tokens, the artificial token [MASK] used in pre-training is absent from labeled data in fine-tuning~\cite{yang2019xlnet}. The discrepancy between pre-training and fine-tuning hinders the model from being adequately optimized on the downstream applications. To alleviate this issue, we propose a gradual fine-tuning by continuing training the models on the set of labeled sets then turning to a specific task. To further boost the model performances on speech-text tasks during fine-tuning, we propose a noisy fine-tuning by perturbing the decoder inputs~\cite{cheng2019robust} in addition to the speech augmentation in the encoder~\cite{park2019specaugment}.

Extensive experimental results on the multilingual CoVoST AST~\cite{wang2021covost}, Voxpopuli ASR~\cite{wang2021voxpopuli} and XTREME benchmarks show that our joint speech-text pre-trained models can achieve competitive results on both speech and text tasks. More specifically, \method{} establishes a new SOTA for models trained on public datasets on CoVoST, with up to 1.9 BLEU points on xx-en and 1.1 BLEU points on en-xx against the previous best results. On Voxpopuli ASR, our model based on a Transformer decoder matches the performance of an \mslam{} model fine-tuned with an RNN-T decoder although the RNN-T decoder is more favorable to ASR tasks. On the multilingual text XTREME, ~\method{} outperforms \mslam{} with 6\% on XNLI, getting closer to the performance of mT5 models of comparable capacity on XNLI and TydiQA. In analyses, we conduct ablation studies to gain further insight into which combination set of supervised datasets in our approach matters the most during the pre-training and fine-tuning. We also vary the noise ratio to investigate the effect of noisy fine-tuning for different speech translation directions.

These results demonstrate that ~\method{} is the first truly multi-modal speech and text model which is capable of performing a wide variety of understanding and generation tasks for speech and text, attaining competitive results with uni-modal text models and vastly improving over speech-only models.

\section{Approach}
\begin{figure}[t]
\centering
\includegraphics[width=0.48\textwidth]{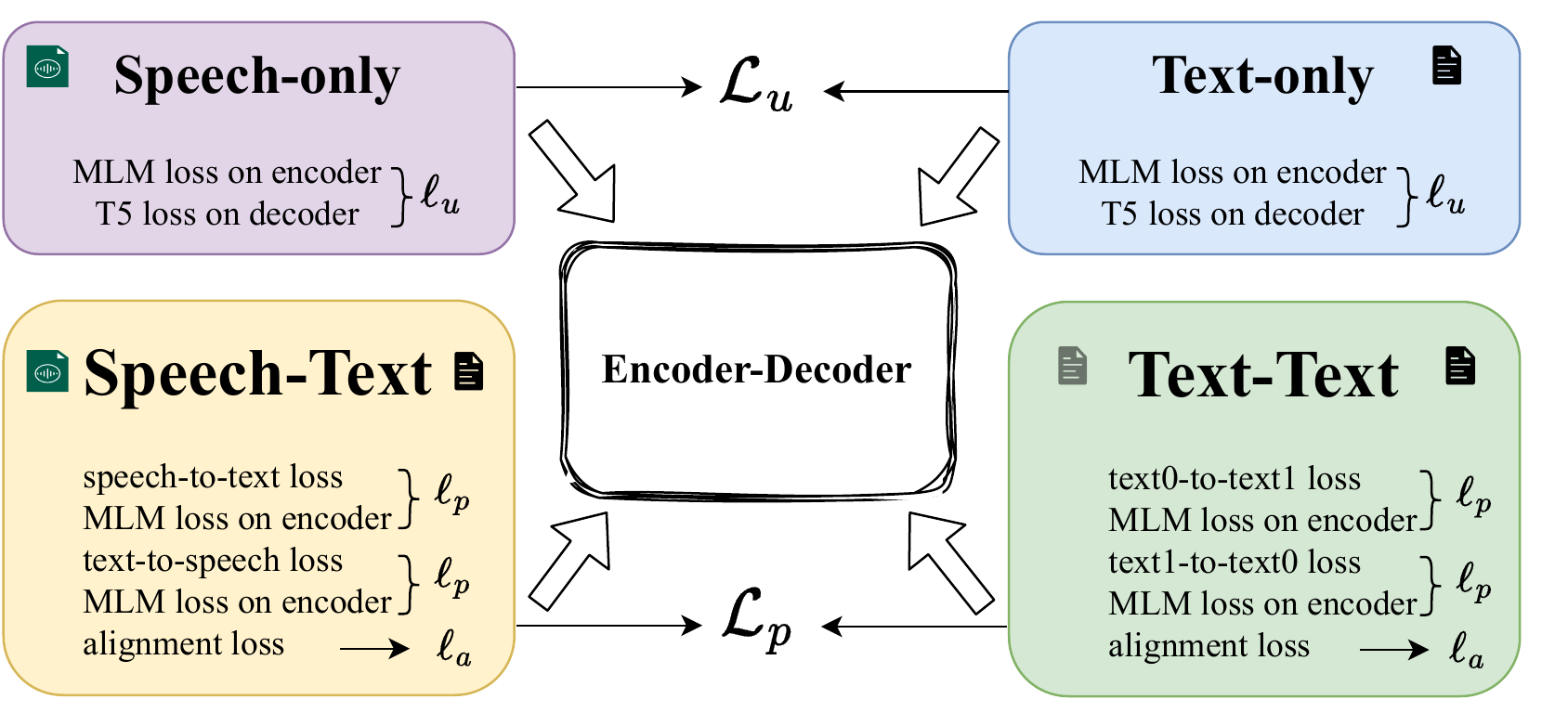} 
\caption{An overview of~\method. A $\ell_u$ loss is used to train speech-only and text-only data by computing masked language modeling (MLM) loss on the encoder and a similar T5 loss on the decoder. The supervised speech-text and text-text data also share the pre-training loss including forward and backward $\ell_{p}$ and an alignment loss $\ell_{a}$ between different languages or modalities. $\ell_{p}$ consists of a translation loss from input to target and a MLM loss on the encoder. Our speech-text models are pre-trained with $\mathcal{L}_u$ on unlabeled data and $\mathcal{L}_p$ on labeled data. In practice, we incorporate an additional CTC loss for ASR.}
\label{figure:approach}
\end{figure}
% some words to give an overview of this method.
We propose a multi-task multilingual pre-training method, ~\method{}, for speech and text, aiming to pre-train speech-text models on arbitrary tasks related to speech and/or text. The speech and text data
can be cast into two types of data, unlabeled data without supervised labels and labeled data usually accompanied with human-annotated labels. As Figure~\ref{figure:approach} shows, we consider four types of data, \ie, speech-only, text-only, speech-text and text-text. The main idea is to unify these training examples into the sequence-to-sequence format and apply similar optimization objectives on the encoder and decoder. The losses on unlabeled data ($\mathcal{L}_{u}$) and labeled data ($\mathcal{L}_{p}$) are combined to pre-train the speech-text models.

\subsection{Model Architecture}
\method{} is based on an encoder-decoder backbone model. For speech inputs, we follow \mslam{}~\cite{bapna:2022} to convert an acoustic feature sequence of 80-dimensional log Mel spectrograms into a sequence of latent speech representations via a CNN block. The CNN block consisting of two 2D-convolutional layers with strides (2, 2) also acts as a sub-sampling mechanism with a 4x reduction in the sequence length dimension. A subsequent linear projection layer is used to map the dimension of the latent speech representations to that of the encoder stack, we denote the speech representations as $\ms$. The text input $\vt$ simply goes through a token embedding layer to be transformed as a sequence of embeddings. To specify the language and modality, we add language and modality embeddings to word embeddings or speech representations $\ms$ in addition to the conventional positional embeddings. The speech and text representations are then fed into a shared multi-modal encoder-decoder model. We prefer a deep encoder with 24 Conformer layers~\cite{gulati:2020} (a 
similar encoder as \mslam{}) and a shallow decoder with 6 Transformer layers~\cite{Vaswani:17}, which favors faster inference while maintaining competitive quality~\cite{kasai:2020}.

\subsection{Speech Tokenization}
The basis of the proposed speech-text pre-training approach is to treat the speech inputs as an additional language, which requires a speech tokenizer to quantize the continuous speech representations $\ms=(\vs_1, \vs_2,..., \vs_N)$ into discrete ids $\vz = (z_1, z_2,...,z_N)$. To this end, each speech representation vector $\vs$ is independently projected into a discrete id $z$ by finding its nearest neighbour in the speech codebook $\mathcal{G}$.
\begin{eqnarray}
z = \argmin_{i} \| \mathcal{G}_{i} - \vs\|.
\end{eqnarray}
In \mslam{}, the parameters of the speech tokenizer are learned from scratch by a contrastive loss~\cite{baevski2020wav2vec} over a speech-only encoder. For simplicity, we directly utilize the pretrained speech tokenizer in mSLAM and keep it constant during our model training.

\subsection{Pre-training Objectives}
In this paper, we have four different training sets related to speech and/or text: a speech-only set $D_{s}$, a text-only set $D_{t}$, a speech-text set $D_{st}$ and a text-text set $D_{tt}$. We want to unify the pre-training losses for unlabeled data and labeled data, which make our pre-training methods easily extensible to more datasets.

\textbf{Losses on unlabeled data} Given an unlabeled training example $\vx = (x_1,x_2,...,x_N)$, we first use it as a source-target pair $(\vx, \vx)$ for the sequence-to-sequence model training. Then we randomly construct a $0/1$ masking vector $\vm$ sampled from a prior distribution. We apply the masking vector $\vm$ to the source $\vx$ by replacing the token $x_i$ with a [MASK] token if $m_i = 1$. The corrupted source $\vx$ is denoted as $\vx^{\vm}$. For the target $\vx$, we employ the complementary masking operation $\neg\vm$ by setting $x_i$ to the [MASK] token if $m_i = 0$ and denote it as $\vx^{\neg\vm}$. Finally, to enable the model to predict the masked source tokens 
on both the encoder and the decoder, the loss $\ell_{u}(\vx^{\vm}, \vx^{\neg\vm};\bm{\theta})$ on the pseudo pair data $(\vx^{\vm}, \vx^{\neg\vm})$ is computed as:
\begin{equation}
\begin{split}
\sum_{m_{i}=1}\log P(x_i|\vx^{\vm};\bm{\theta_{e}}) + \sum_{m_{i}=1} \log P(x_i|\vx^{\neg\vm}_{< i},\vx^{\vm}; \bm{\theta}), \label{loss:unpaired_data}
\end{split}
\end{equation}
where the parameter set %of the encoder-decoder model
$\bm{\theta} = \{\bm{\theta_{e}},\bm{\theta_{d}}\} $ is split into two parts, \ie, $\bm{\theta_{e}}$ for the encoder and $\bm{\theta_{d}}$ for the decoder.

It is natural to use the above loss to pre-train the model on the text-only data set $D_{t}$. For speech, we take the representations $\ms$ extracted by the CNN block as inputs and their corresponding discrete ids $\vz$ quantized by the speech tokenizer as targets. Likewise, we can also mask out the speech representations by substituting the embedding of the [MASK] token for $\ms_i$ if $m_i = 1$  and applying $\neg\vm$ to $\vz$, 
because they have identical lengths. The loss on the unlabeled speech-only ($D_{s}$) and text-only ($D_{t}$) sets is:
\begin{equation}
\begin{split}
\mathcal{L}_{u}(\bm{\theta}) =& \mathbb{E}_{\vs \in D_{s}} \mathbb{E}_{\vm} [\ell_{u}(\ms^{\vm}, \vz^{\neg\vm};\bm{\theta})] + \\
&\mathbb{E}_{\vt \in D_{t}} \mathbb{E}_{\vm} [\ell_{u}(\vt^{\vm}, \vt^{\neg\vm};\bm{\theta})],
\end{split}
\end{equation}
where we allow the sequence of speech representations $\ms$ to be passed into $\ell_u$ for convenience.

\textbf{Losses on labeled data} For a labeled example $(\vx, \vy)$ where $\vx = (x_1,x_2,...,x_N)$ and $\vy = (y_1,y_2,...,y_M)$, we employ its forward and backward sequence-to-sequence loss, \ie, $P(\vy|\vx;\bm{\theta})$ and $P(\vx|\vy;\bm{\theta})$. Since the labeled training data is usually not abundant compared to the unlabeled data, we introduce a similar mask operation $\vm$ for the source part of the labeled data to avoid overfitting. Meanwhile, the reconstruction loss in the encoder is 
also applied to enhance the representation learning for the deeper encoder. Thus the forward sequence-to-sequence loss $\ell_{p}(\vx^{\vm}, \vy;\bm{\theta})$ on the paired data $(\vx^{\vm}, \vy)$ is calculated as:
\begin{eqnarray}
\sum_{m_{i}=1}\log P(x_i|\vx^{\vm};\bm{\theta_{e}}) + \sum_{i} \log P(y_i|y_{< i},\vx^{\vm}; \bm{\theta}). \label{loss:seqtoseq}
\end{eqnarray}
To better align learned representations between different languages and modalities, except for the fully shared encoder-decoder model to implicitly encourage the alignment, we also introduce an explicit alignment loss on the encoder and decoder. Given the paired data $(\vx, \vy)$, they are concatenated into a new sentence $[\vx,\vy]$ where $[\cdot,\cdot]$ stands for the concatenation along the sequence dimension. Similar to computing the loss on unlabeled data, a randomly sampled mask vector $\vm$ and its complementary mask $\neg\vm$ are manipulated over $[\vx,\vy]$, which results in a masked pair $([\vx,\vy]^{\vm}, [\vx,\vy]^{\neg\vm})$. We compute the pre-training loss over this masked pair through Eq. (\ref{loss:unpaired_data}). In practice, we observe that the individual predictions for masked tokens of either $\vx^{\neg\vm}$ or $\vy^{\neg\vm}$ on the decoder performs better, particularly, it exhibits stronger stability during pre-training. Therefore, we compute the alignment loss over $\ell_{a}([\vx,\vy]^{\vm}, [\vx,\vy]^{\neg\vm})$ as: 
\begin{eqnarray}
\sum_{m_{i}=1}\log P([\vx,\vy]|[\vx,\vy]^{\vm};\bm{\theta_{e}}) \nonumber \\  
+ \sum_{m_{i}=1} \log P(x_i|\vx^{\neg\vm}_{< i},[\vx,\vy]^{\vm};\bm{\theta}) \nonumber \\
+\sum_{m_{i}=1} \log P(y_i|\vy^{\neg\vm}_{< i},[\vx,\vy]^{\vm};\bm{\theta}).
\label{loss:align}
\end{eqnarray}
To sum up, the losses on the %labeled
speech-text ($D_{st}$) and text-text ($D_{tt}$) sets involve the forward and backward sequence-to-sequence losses and an alignment loss for each example:
\begin{equation}
\begin{split}
\mathcal{L}_{p}(\bm{\theta}) &= \mathbb{E}_{(\vs, \vt) \in D_{st}} \biggl\{ \mathbb{E}_{\vm} [\ell_{p}(\ms^{\vm}, \vt^{\neg\vm};\bm{\theta})] + \\
&\mathbb{E}_{\vm} [\ell_{p}(\vt^{\vm}, \vz^{\neg\vm};\bm{\theta})]  + \mathbb{E}_{\vm} [\ell_{a}([\ms,\vt]^{\vm}, [\vz, \vt]^{\neg\vm}) ] \biggr\}  \\
&+ \mathbb{E}_{(\vv, \vt) \in D_{tt}} \biggl\{ \mathbb{E}_{\vm} [\ell_{p}(\vv^{\vm}, \vt^{\neg\vm};\bm{\theta})] +  \\
&\mathbb{E}_{\vm} [\ell_{p}(\vt^{\vm}, \vv^{\neg\vm};\bm{\theta})] + \mathbb{E}_{\vm} [\ell_{a}([\vv,\vt]^{\vm}, [\vv, \vt]^{\neg\vm}) ] \biggr\}
\end{split}
\end{equation}
where the sequence of speech representations $\ms$ and the sequence of text ids $\vt$ are allowed to be concatenated in the formula just for convenience.
In addition to the above conventional loss across all the paired training sets, we follow mSLAM~\cite{bapna:2022} to leverage the CTC loss~\cite{graves2006connectionist,graves2014towards} to enforce the alignment of the encoder representations between speech and text, which is only activated on ASR data.

\subsection{Fine-tuning}
The fine-tuning method is crucial to unlock the capability of pre-trained models.
In this section, We  explain the process of fine-tuning our multilingual speech-text models for downstream tasks related to both speech and text.

\textbf{Direct fine-tuning}
A common way of adapting a pre-trained model to a specific downstream task is to continue training the model exclusively on labeled data from that task, usually in combination with a relatively smaller learning rate. We use this direct fine-tuning for those tasks that are not included in our labeled data sets, \eg, text classification.

\textbf{Gradual fine-tuning}
To mitigate the discrepancy between pre-training and fine-tuning due to the artificial [MASK] tokens in pre-training~\cite{yang2019xlnet}, we propose a two-stage gradual fine-tuning method. At the first stage, as we incorporate labeled datasets from the downstream tasks of our interest during the pre-training phase, we keep training the models by using the sequence-to-sequence loss (Eq.~(\ref{loss:seqtoseq})) on the paired data but disabling [MASK] tokens. Only $P(\vy|\vx;\bm{\theta})$ for a pair $(\vx, \vy)$ is used after the mask operation is eliminated.
Since we fine-tune the model over the combination set of multiple tasks with numerous languages, we call this fine-tuning method at this stage as {\em multi-task multilingual fine-tuning}. At the second stage, we further continue fine-tuning the model on one of tasks from the first-stage combination set.

\textbf{Noisy fine-tuning} When fine-tuning speech-text tasks, an augmentation method directly acting on speech spectrogram is exploited to prevent the overfit on the limited supervised set~\cite{park2019specaugment}. However, the perturbations in the source introduced by the augmentation method tend to affect the decoder predictions in which errors may be accumulated and amplified at the later steps. To defend these errors from the decoder, we follow~\cite{cheng2019robust} to add some noise to decoder inputs. More specifically, we randomly replace some tokens in the decoder inputs with their synonym tokens measured by the word embeddings.

\section{Experiments}
\covostmanytomany
\subsection{Setup\protect\footnote{More details for the setup can be found in the appendix.}}
\textbf{Data} Following \mslam{}~\cite{bapna:2022}, we use the same unlabeled speech data of approximately 429k hours in 51 languages. The mC4 dataset spanning 101 languages is used as unlabeled text data. ASR data come from VoxPopuli, MLS, Babel, CoVoST and FLEURS. We only have two sources for AST data, CoVoST and FLEURS. We collect MT data from WMT and TED.

\textbf{Model} We use an identical Conformer layer from SLAM~\cite{bapna:2021} and mSLAM~\cite{bapna:2022}. The Transformer layers in the decoder share a similar setting as Conformer layer. The Adam optimizer~\cite{kingma2014adam} is applied to pre-traing while AdamW~\cite{loshchilov2017decoupled} is used for fine-tuning.

\textbf{Pre-training} The batch sizes per TPU for speech-only, text-only, AST, ASR and MT data are 4, 8, 1, 1 and 1. We mask approximately $50\%$ of the speech frames with spans of length up to 10~\cite{chung2021w2v}. However, for text inputs, we mask a continuous span of around $50\%$ of words except for MT tasks where the mask ratio is $25\%$. The loss coefficients related to speech-only and text-only data are set to $1$.
The loss coefficients for the text to speech and alignment tasks  are $0.1$ while speech to text tasks need a slightly higher loss coefficient $0.3$ for the decoder loss. We pre-train two sets of speech-text models in which two different text vocabularies are used, \ie, a character-level model (\method{}-char) of $4096$ chars and a spm-level model (\method{}-spm) of $64k$ word pieces. These two models run on $256$ TPUv4 chips for $1.5M$ steps.

\textbf{Fine-tuning} We fine-tune our pre-trained models on CoVoST-2 multilingual speech translation~\cite{wang2021covost}, VoxPopuli multilingual speech recognition~\cite{wang2021voxpopuli}, and XTREME multilingual text understanding~\cite{hu2020xtreme} benchmarks. We report the detokenized BLEU scores calculated by the SacreBLEU script~\cite{post2018call}.
For each fine-tuning tasks, we use grid search to tune the hyperparameters including batch sizes per TPU over $\{2, 4, 8\}$, learning rates over $\{0.5, 1, 2, 3, 5\}$, dropout ratios for encoder inputs and Transformer decoder over $\{0.1, 0.3\}$, warm-up steps over $\{4k, 8k, 16k\}$. Generally, we observe that speech-related and text-related tasks are not very sensitive to the batch size so we 
use $8$. Speech-related tasks prefer a larger learning rate of $5$ while text-related tasks needs a smaller one of $1$ or $0.5$. The warm-up steps are universally set to $16k$. The pre-trained spm-level model is in favor of a larger dropout of $0.3$. In our multi-task multilingual fine-tuning experiments, the training examples of AST, ASR and MT for a batch is set to $4$, $2$ and $2$. For AST, ASR and MT, we randomly incorporate synonym noises into decoder inputs, the noise ratio is set to $0.06$. All of fine-tuning experiments are conducted on $64$ TPUv4 chips. Except for the multi-task multilingual 
fine-tuning experiments in which we select a maximum fine-tuning step of $300k$ and report results from the last checkpoint, we pick the best model based on validation sets.

\covostentoxx
\asrvp

\subsection{Multilingual Speech Translation}
Table~\ref{table:covost_xx2xx} shows BLEU scores on the CoVoST 2 dataset by fine-tuning the pre-trained models on English to non-English (en-xx) and non-English to English (xx-en) language pairs. We try three fine-tuning setups: (1) direct multilingual fine-tuning with xx-en or en-xx language pairs; (2) multi-task multilingual fine-tuning with all of available language pairs from AST, ASR and MT; (3) gradual fine-tuning by further training the model on xx-en or en-xx language pairs. We observe that direct fine-tuning with only AST data can already obtain better performance against XLS-R and $0.6$B mSLAM models (up to $+1.5$ BLEU points). When multi-task multilingual fine-tuning is applied, the model can achieve better results on xx-en but lower scores on en-xx. As this model can be used for multiple tasks, not just limited to AST, it is reasonable that the good performance on en-xx can not be kept. We believe that a model with larger capacity is able to improve the results on en-xx. However, in terms of average scores on all of language pairs in AST, the model still makes some improvements (up to $+0.7$ BLEU points). The best models are delivered by using gradual fine-tuning on \method{}-char and \method{}-spm. We establish new SOTA results on xx-en with $+1.9$ BLEU gains compared to Maestro~\cite{chen2022maestro}. Meanwhile, we also notice that \method{}-spm reaps more benefits from multi-task multilingual fine-tuning, particularly on en-xx. We speculate that smaller granularity of character-level vocabulary is conductible to language transfer because of more characters shared across different languages and the domination of English tokens during the SPM vocabulary creation. Thus \method{}-char already gets good results without very advanced fine-tuning techniques. The comparison between direct fine-tuning and gradual fine-tuning clearly show that the multi-stage fine-tuning is indispensable for our models to get better results.

Additionally, we compare the results between Whisper~\cite{radford2022robust}, which evaluates the benchmark in a zero-shot manner, and our proposed method~\method{}. The superior performance of Whisper on xx-en suggests that the straightforward scaling of weakly supervised pre-training holds great potential. However, we think the comparison isn't exactly fair since Whisper~\cite{radford2022robust} is using an order of magnitude larger proprietary dataset and also evaluating out of domain. The Whisper~\cite{radford2022robust} paper highlights the effectiveness of weakly supervised pre-training for building a general-purpose ASR + AST system, whereas our method focuses more on leveraging unlabeled speech and text pre-training data, and multilingual multitask supervised data towards learning a single model for speech and text understanding. Morever, the Whisper model is much larger than our~\method{} model (1.6B vs. 0.6B).

Table~\ref{table:covost_en2xx} shows AST results on four English to Non-English (en-xx) directions from CoVoST 2 by following the identical setting in baseline methods. To make a fair comparison, at the second stage in the gradual fine-tuning, we fine-tune the model only with a single language pair. Our models outperform the previous best SpeechLM~\cite{zhang2022speechlm} with up to $+1.1$ BLEU points and \method{}-char still performs slightly better than \method{}-spm on en-xx.

To sum up, we have the following findings from these two tables: (1) gradual fine-tuning is tremendously helpful to improve model performance; (2) \method{}-spam model gains much more from gradual fine-tuning; (3) \method{}-spam is in favor of xx-en translation directions while \method{}-char performs much better on en-xx.

\xtremexnli
\xtremeqa

\subsection{Multilingual Speech Recognition}
We present ASR results on the multilingual VoxPopuli dataset in Table~\ref{table:asrvp}. There are three different fine-tuning setups, multilingual fine-tuning only with ASR, multi-task multilingual fine-tuning and gradual fine-tuning from multi-task multilingual fine-tuning to ASR-only multilingual fine-tuning. If we directly evaluate the models fine-tuned with multi-task multilingual fine-tuning, we can find both of them can not achieve reasonable numbers in this benchmark. It might be because AST data dominates the multi-task multilingual fine-tuning and deteriorates the monotonic alignments between encoder and decoder but ASR tasks are very sensitive to heterogeneous data.

In the other two setups, \method{}-spm performs better than \method{}-char, particularly, \method{}-spm benefits more from gradual fine-tuning with around +0.5 WER gains. It indicates that speech-text model based on the SPM vocabulary has more potential of attaining better results if being elaborately fine-tuned with observing more paired data.
Before multi-task multilingual fine-tuning, our best model (\method{}-spm) can only beat XLS-R. However, the exploitation of multi-task multilingual fine-tuning makes our best model achieve similar performance against mSLAM although it still lags behind Maestro. That is because Transformer~\cite{Vaswani:17} rather than RNN Transducer~\cite{graves2012sequence} is applied as a decoder in our speech-text ASR model, but it has demonstrated the effectiveness of jointly pre-training encoder and decoder for learning better speech-text representations to even dispense with Transducer decoder. The Whisper model also adopts the Transformer model with a larger model capacity and benefits from a larger amount of pre-training data. However, it is worth mentioning that the Whisper model is not particularly proficient in ASR tasks and, in fact, it demonstrates significantly poorer performance compared to our \method{} approach.

\covostablation

\subsection{Multilingual Text Understanding}
We also investigate the capability of our speech-text models with respect to text understanding on two representative evaluation tasks from XTREME~\cite{hu2020xtreme}, from which we pick up two representative evaluation tasks, XNLI classification and TyDiQA question answering tasks.

\textbf{XNLI classification task} As shown in Table~\ref{table:xnli}, in the zero-shot setting, our model underperforms the mono-modal multilingual model of the similar size, mT5-Base (0.6B). We think that capacity dilution results in the degeneration of our speech-text models against text-only modeling. The best speech-text joint model \method{}-spm can achieve better results than mT5-Small (0.3B) which is roughly half the model size of \method{}-spm. We believe the increased model capacity can compensate for the decease of \method{} against mT5. When compared with \mslam{} models, \method{} consistently performs significantly better than the \mslam{}($0.6$B) model ($66.4$ vs.~$58.9$) and comparable with the \mslam{} model ($2$B). More specifically, \method{} obtains notable improvements on non-European languages. The comparison between spm-level and character-level models indicates that spm-level tokens are better to capture the text meaning. Similar findings can be observed when moving to the {\em Translate-Train-All} setting.

\textbf{TyDiQA task} Table~\ref{table:tydiqa} shows F1/EM results on a multilingual question answering task, TyDiQA. In the zero-shot setting, similar to classification results in Table~\ref{table:xnli}, \method{}-spm maintains better performance than \method{}-char. However, our \method{} models can not even surpass mT5-base although we achieve significantly better results than mT5-base on XNLI. We analyze the breakdown results on each language pair attached in the appendix. We find our models are able to deliver good results on English but relatively worse results on non-English languages. More specifically, our models nearly approach $0$ on Bengali language. After looking into the model outputs, we find the model can not properly output the correct tokens in non-English languages. We ascribe this issue to the language embeddings which do not specialize in generation outside the fine-tuning languages. Likewise, we evaluate our models on the {\em Translate-Train-All} setting. As expected, our models are better than mT5-Small but worse than mT5-Base.

\subsection{Analysis}

\textbf{Effect of paired data} To study the effect of speech-text and text-text labeled data, we conduct extensive experiments by using different combination sets of AST, ASR and MT data during pre-training and fine-tuning. In Row 1-4, only AST data is enabled in fine-tuning.
%for fine-tuning pre-trained models.
The best model comes from Row 4 which digests all available speech-text data. It is interesting that removing ASR data improves the model performance on en-xx directions. We speculate that translation data (AST and MT) is important to learn the alignment between encoder and decoder but ASR with very strong monotonic alignment hurts the alignment, particularly for en-xx in which a specific non-English translation data is not as abundant as English. When applying multi-task multilingual fine-tuning, we find the model with all data in pre-training does not perform the best. Row 6 without MT involved in pre-training takes the lead in the AST results. It is probably because text occupies model capacity in encoder pre-training but the effect of MT data is made up during the fine-tuning stage.  These comparisons suggest that multi-task pre-training is beneficial to learning general speech-text representations if we do not have any assumption on downstream tasks. As we want to evaluate our models on both speech and text tasks, we incorporate all of available labeled data related to speech and text in our model pre-training.

\begin{figure}[t]
\centering
\includegraphics[width=0.48\textwidth]{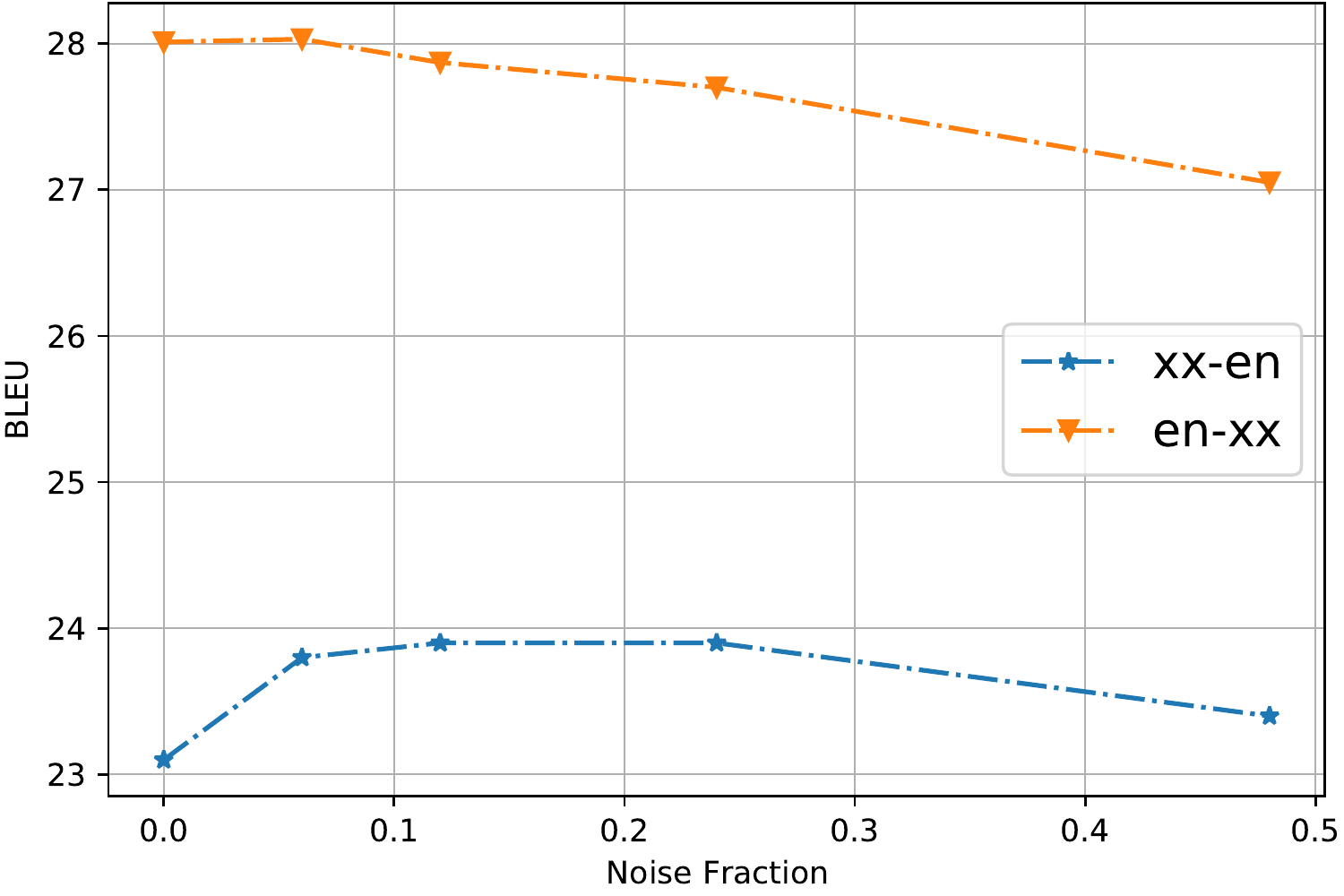} 
\caption{Effect of noisy fine-tuning when changing the noise ratio.}
\label{figure:robustness}
\end{figure}

\textbf{Effect of noisy fine-tuning} We randomly replace decoder inputs with their synonym tokens in noisy fine-tuning. Figure~\ref{figure:robustness} shows the sensitive study results on the noise ratio. When using the~\method{}-char model on xx-en, the BLEU score has a drastic change when the noise ratio increases from $0$ to $0.06$, then reaches the plateau, finally drops a lot as it increases to $0.48$. However, for en-xx, the noisy fine-tuning has subtle improvements when using a non-zero  small noise ratio ($0.06$). If increasing it further, it hurts the model performance severely. The improvement differences between xx-en and en-xx imply that the noise on the decoder is more favourable to the same or similar languages (xx-en) rather than a set of diverse languages (en-xx)~\cite{cheng2022multilingual}. In addition, the word embeddings can not accurately measure the similarities between different languages.

\section{Related Work}

\textbf{Speech-text Pre-training} 
Pre-training methods have dominated research and industry fields due to their superior capabilities of exploiting unlabeled data. Particularly,
in NLP and speech. A lot of representatives, such as BERT~\cite{devlin2018bert}, XLNET~\cite{yang2019xlnet}, T5~\cite{raffel2020exploring}, MASS~\cite{song2019mass}, wav2vec~\cite{baevski2020wav2vec}, Hubert~\cite{hsu2021hubert} and so on, come out to improve mono-modal model performance. In recent days, the research community has started to move towards speech-text joint training, aiming to learn the shared representation space between speech and text, which can be roughly categorized 
into, encoder-only pre-training~\cite{bapna:2021,bapna:2022,zhang2022speechlm}, encoder-decoder pre-training~\cite{ao2021speecht5,lakhotia2021generative,chen2022maestro,sainath2022joist,Zhou2022MMSpeechMM,zhang2022speechut,tang2022unified,popuri2022enhanced,radford2022robust}.~\method{} adopts an encoder-decoder backbone model by minimizing the utilization of modality-specific blocks only with a CNN used to extract speech representations, 
which dramatically simplifies the cross-modal model architecture and also enforces the representation sharing between different languages and modalities. Among them, the most related Maestro~\cite{chen2022maestro} also incorporates ASR, AST and MT data into their method, however, their model training has to rely on a pre-trained \mslam{} as initialization and applies a duration model to over-sample the text which can not be activated during fine-tuning. In contrast,~\method{} pre-trains the model from scratch which can be applied in the downstream tasks without wasting any parameter. We also verify our models on text-related benchmarks while they just focus on speech tasks. In addition, a text-to-speech loss is also introduced in pre-training which endows our model with the ability of speech generation. We leave it as the future exploration.

\textbf{Multilingual Pre-training} The great success of multilingual text pre-training like mBERT~\cite{devlin2018multilingualbert}, XLM-R~\cite{conneau2019cross} and mT5~\cite{xue2020mt5} incentivizes the speech research to move toward multilingual modeling and pre-training~\cite{conneau2020unsupervised,babu2021xls,bapna:2022,chen2022maestro,radford2022robust}, which benefits from the cross-lingual transfer for learning joint representations across massive amounts of data across multiple languages~\cite{conneau2019unsupervised,hu2020xtreme}. Our approach is inspired by this research direction by involving multilingual speech and text spanning over 100 languages.

\textbf{Multi-task Learning} Multi-task learning is an effective approach that utilizes the training signals of related tasks to enhance the generalization performance of a model~\cite{Caruana1997}. This technique has been successfully applied to improve various speech-related tasks, as evidenced by previous studies~\cite{weiss2017sequence,tang2021general,tang2022unified,chen2022maestro,bapna:2022}. In our paper, we adopt a unified approach by combining speech and text-related tasks into a single sequence-to-sequence model during the pre-training stage. Our aim is to leverage the training signals from diverse speech and/or text-related tasks, encompassing speech-only, text-only, ASR, AST, MT, and potentially TTS, in order to maximize the benefits of multi-task learning.

\section{Conclusion}

We have presented~\method{} for speech and text joint models based on a fully encoder-decoder model. Our pre-training models span more than 100 languages in speech and text and involve unlabeled data and labeled data from speech/text-only data, ASR, AST to MT. We introduce two training objectives to unify the unlabeled and labeled data in pre-training, and gradual fine-tuning and noisy fine-tuning to improve the model performance on downstream tasks. Extensive experiments on multilingual benchmarks show that our pre-training models can achieve very strong results with new SOTA on CoVoST and comparable performance against~\mslam{} on VoxPopuli, and narrow the gap between speech-text models and text-only models on text tasks. 

\section{Limitations and Future Work}

Based on our extensive experiments, we have identified several limitations in this paper, which we believe open up potential avenues for future research and development.

\begin{enumerate}
    \item While we pre-train our model on the text-to-speech task, we have not evaluated our approach on speech generation benchmarks. It would be beneficial to include them in our evaluation process to provide a more comprehensive assessment of the model's performance across different modalities.
    \item Our model is limited to only 100 languages available in academic datasets unlike models trained on proprietary datasets, \eg USM~\cite{zhang2023google}. We plan to scale up our model beyond 100 language.
    \item We have not integrated speech to speech tasks into our pre-training framework to further expand the capabilities of the model and explore the potential benefits of jointly training on these tasks~\cite{popuri2022enhanced}.
    \item Our pre-trained model is limited in zero-shot speech translation and recognition, as well as zero-shot text generation task. It is worthwhile improving model's alignment transfer between unseen language pairs and modalities. One of possible directions is to switch to decoder-only models~\cite{anil2023palm}.
    \item Our proposed approach still relies on speech representations as inputs in encoder , rather than solely relying on token-to-token transformations.
\end{enumerate}

\section*{Acknowledgements}
The authors would like to thank anonymous reviewers for
insightful comments, which greatly contributed to the improvement of this paper. Special thanks are given to Yuan Cao and Zhehuai Chen for their invaluable discussions and contributions during the model training. %Their input and expertise have been instrumental in shaping the development of this research.

\balance
\bibliography{icml2023}
\bibliographystyle{icml2023}

%%%%%%%%%%%%%%%%%%%%%%%%%%%%%%%%%%%%%%%%%%%%%%%%%%%%%%%%%%%%%%%%%%%%%%%%%%%%%%%
%%%%%%%%%%%%%%%%%%%%%%%%%%%%%%%%%%%%%%%%%%%%%%%%%%%%%%%%%%%%%%%%%%%%%%%%%%%%%%%
% APPENDIX
%%%%%%%%%%%%%%%%%%%%%%%%%%%%%%%%%%%%%%%%%%%%%%%%%%%%%%%%%%%%%%%%%%%%%%%%%%%%%%%
%%%%%%%%%%%%%%%%%%%%%%%%%%%%%%%%%%%%%%%%%%%%%%%%%%%%%%%%%%%%%%%%%%%%%%%%%%%%%%%
\newpage
\appendix
\onecolumn
\section{Setup}
\subsection{Data}
\textbf{Speech-only Data} Following \mslam{}~\cite{bapna:2022}, we use the same unlabeled speech data of approximately 429k hours in 51 languages from VoxPopuli~\cite{wang2021voxpopuli}, Common Voice~\cite{ardila2019common}, MLS~\cite{pratap2020mls} and Babel~\cite{Gales2014SpeechRA}.

\textbf{Text-only Data} The mC4 dataset~\cite{xue2020mt5} spanning 101 languages is used as unlabeled text data by adopting a temperature-based sampling to over-sample the low-resource languages where temperature is $3$.

\textbf{Speech-Text Data} 
We use ASR data from VoxPopuli of approximately 1.3k hours with 14 languages, MLS of 80 hours with 8 languages, Babel of 1k hours  with 17 languages, CoVoST of 2.9k hours with 22 languages~\cite{wang2021covost}, FLEURS of 1.4k~\cite{conneau2022fleurs} hours with 101 languages. We only have two sources for AST data, CoVoST of 9.5k hours spanning 22 languages and FLEURS of 1.4k hours spanning 101 languages.

\textbf{Text-Text Data} The paired text-text data comes from WMT and TED translation tasks, which are identical as the MT sets in \mslam{}\cite{bapna:2022}. More specifically, we collect MT data from WMT and TED which has the similar language coverage as CoVoST. We pair  WMT20~\cite{barrault-etal-2020-findings} for ja, ta,  
WMT19~\cite{barrault-etal-2019-findings} for de, ru, zh, 
WMT18~\cite{bojar-etal-2018-findings} for et, tr,  
WMT17~\cite{bojar-etal-2017-findings} for lv, 
WMT15~\cite{bojar-etal-2015-findings} for fr, 
WMT13~\cite{bojar-etal-2013-findings} for es, and 
TED59~\cite{qi-etal-2018-pre} for ar, fa, id, it, mn, nl, pt, sl, sv, leaving ca and cy unpaired.  Because the language distribution in this combination set is highly skewed, we also apply the similar temperature-based data sampling with temperature as $2$. 
\subsection{Model and Hyperparameters}
\textbf{Model setup} We use an identical Conformer layer from SLAM~\cite{bapna:2021} and mSLAM~\cite{bapna:2022}, in which the model dimension is $1024$, feedforward hidden dimension is $4096$, convolution kernel size is $5$ and the number of attention heads is 8. The Transformer layers in the decoder share the same setting as Conformer layer in terms of model dimension, hidden dimension and attention heads but we set dropout to $0.1$ for the Transformer layers rather than the default $0$ in the Conformer layers. We use the same learning schedule as Transformer during pre-training and fine-tuning but warmup steps are set to $40k$ and $16k$ respectively. The Adam optimizer~\cite{kingma2014adam} is applied to pre-traing with learning rate as 3 while AdamW~\cite{loshchilov2017decoupled} is used as fine-tuning optimizer with weight decay rate as $0.01$.

\textbf{Pre-training setup}
For text masking, we follow~\cite{devlin2018bert} by replacing the masked tokens with (1) the [MASK] token 80\% of the time (2) a random token 10\% of the time (3) the unchanged i-th token 10\% of the time. For speech masking, if a token needs to be masked, we just simply replace it with the [MASK] token.
\covostperlanguage
\covostperlanguageentoxx
\vpperlanguage
\xnliperlanguage
\tydiqaperlanguage

\end{document}